\documentclass[runningheads]{llncs}
\usepackage{graphicx}
\usepackage{subfig}

\begin{document}
\title{ Emotion Recognition with Facial Attention and Objective Activation Functions}
\titlerunning{FER with Attention and Objective Activation Functions}
\author{Andrzej Miskow\inst{1}\orcidID{0000-0002-9666-0411} \and
Abdulrahman Altahhan\inst{2,3}\orcidID{0000-0003-1133-7744} }
\authorrunning{A. Miskow et al.}
\institute{School of Computing, University of Leeds.}
\maketitle 
\begin{abstract}

In this paper, we study the effect of introducing channel and spatial attention mechanisms, namely  SEN-Net, ECA-Net, and CBAM, to existing CNN vision-based models such as VGGNet, ResNet, and ResNetV2 to perform the Facial Emotion Recognition task. We show that not only attention can significantly improve the performance of these models but also that combining them with a different activation function can further help increase the performance of these models.

\end{abstract}
\section{Introduction}

The most recent breakthrough in emotion recognition is the idea of using attention to improve the accuracy of the deep learning model. The methodology behind visual-attention-based models was inspired by how humans inspect a scene at first glance. \cite{Desimone1995} has found that humans retrieve parts of the scene or objects sequentially to find the relevant information. Since neural networks attempt to mimic how the human brain works to complete the desired task, various methods were developed to imitate human attention. The discovery of these attention mechanisms helped improve the accuracy of emotion recognition models. 
In this work, we aim to discover the effect of introducing an attention mechanism to existing deep learning models to recognise facial expressions and how their performance can be further boosted via simple but effective changes to their architectures. Additionally, the new architectures will be further improved by modifying their activation functions from ReLU to ELU activation functions to solve the issue of bias shift.
The paper proceeds as follows. In the next section, we present related work, while in section 3, we show the methodology, and in section 4, we show the results.

\section{Attention}

When processing a complex visual scene, human vision does not process the entire image at once. Instead, we tend to only focus on a subset of the image while ignoring the rest to speed up the visual analysis process. This process of selecting a subset of the input, and ignoring the rest, is referred to as attention \cite{Connor2004}. Attention can be divided into two independent categories \cite{Pinto2013}: bottom-up unconscious (implicit) attention, referred to as saliency-based attention that operates on raw sensory input, and top-down conscious (explicit) attention, which refers to the deliberate allocation of attention to certain features. 

\subsection{Attention in Facial Expression Recognition Context}
In computer vision, attention mechanisms can be treated as a selection process that weighs an input dimension and its features according to their importance to the task. Each dimension defines a different input domain that contributes to the task to a different extent. Furthermore, each portion of the input domain has varied importance to the task.

From a deep learning perspective, attention can be infused into a CNN model by further distinguishing higher-level features from low-level features and assigning higher weights to crucial features, i.e., by attracting the model's attention to these features \cite{Hu2018}. From this perspective, attention mechanisms can be divided into three distinct categories;  channel, spatial, and temporal attention. Additionally, these categories can be combined to form other hybrid attention mechanisms, namely: channel \& spatial attention and spatial \& temporal attention. Temporal attention will not be discussed as the work focuses on recognition from static images, not sequential data.

Since the introduction of attention modules and their easy integration with CNN's models, researchers have switched their focus to using CNN classification with attention for FER applications. In \cite{Cohn1995}, authors proposed using attention for FER tasks. They implemented spatial attention with a CNN and improved the model's performance by focusing less on irrelevant parts of the image and training the model on "where" to find the needed information. Additionally, \cite{Li2021} has explored using channel attention which defines "what" to look for in the model by placing a higher value on more informative features. The combination of spatial and channel attention for FER applications is envisaged to achieve state-of-the-art results. 

\section{Methodology}
We use three different CNN image processing models as our base model and add attention to them to boost their performances. The models that we use are VGGNet, Resnet, and ResnetV2. These models are considered a good fit for our problem due to their resilience to noise and ability to deal with degradation and vanishing gradient problems. Each one of these models has its strengths and weaknesses, and we want to study what happens when we add attention to them in the context of FER. 

In addition, we vary the depth of these architectures to study the effect of different attention mechanisms on the depth of the architecture and whether they aggravate or alleviate some of the issues associated with the depth of the architecture. Furthermore, to make our study more comprehensive, we also study the effect of the activation function on these architectures when integrated with each attention mechanism.

This section starts by discussing the preprocessing stage that we adopted. Then we move to the activation functions and show a preliminary comparative study for a lab-based FER dataset, the CK+. We then discuss the different attention modules and conclude the section by conducting preliminary experiments on the reduction rate of the attention modules, again using the CK+ dataset. This section is followed by full-fledged experimental results that compare all the different architecture performances on the more challenging FER2013 dataset.
\subsection{Face Detection and Pre-processing} 
We start by detecting the face in the image and removing the insignificant background pixels.
Without this step, unwanted features in the image may be extracted and classified along with important information resulting in errors.
Facial detection can be achieved using standard object detection methods. This paper uses a state-of-the-art facial detector built on top of the YOLO  framework \cite{Qi2021}. YOLO was chosen due to its efficient one-stage object detection capability comparable to the performances of two-stage detectors while offering significantly better computational performance \cite{Bochkovskiy2020}.

 The default \textit{yolov5s} weights were chosen due to their high performance and accuracy after experimenting with different weights on a subset of the dataset. More importantly, the original YOLO architecture was modified to ensure the output images had a fixed image size of $80 \times 80$ pixels. Since faces bounding boxes can have different proportions, cropped faces must be re-sized, so they all have the same size. This stage can be considered an external attention layer for our model.

\subsection{Activation Functions}
 \subsubsection{ReLU} activation function has helped to solve the vanishing gradient problem, and hence it was utilised by the architectures discussed earlier. This is because the gradients of the ReLU activation follow the identity function for positive arguments and zero otherwise, meaning that large gradient values are still used, and negative values are discarded. On the other hand, since ReLU is non-negative, it has a mean activation larger than zero. As a result, neurons with a non-zero mean activation act as a bias for the next layer causing a bias shift for the next layer. The shift in bias causes weight variance, leading to the activation function being locked to negative values, and the affected neuron can no longer contribute to the network learning. Consequently, two activation functions have been proposed that tackle the problem of bias shift differently while also solving the vanishing gradient problem.
\subsubsection{ELU} function was proposed that allows negative gradient values, resulting in the mean of the unit activations being closer to zero than ReLU. Like ReLU, ELU applies the identity function for positive values, whereas it utilises the exponential function if the input is negative. For this reason, ELU achieves faster learning, and significantly better generalization performance than ReLU on networks with more than five layers \cite{Clevert2016}.

\subsubsection{SELU}  function \cite{Klambauer2017} was proposed to solve the issue of bias-shift through self-normalization. Through this property, activations automatically converge to a zero mean and unit variance. This convergence property makes SELU ideal for networks with many layers and further improves the ReLU activation function.

\begin{table}\caption{Performance of the activation functions on the CK+ dataset with ResNet-50}\label{table1}
\centering
\begin{tabular}{|c|c|}
\hline
Activation Function & Accuracy \\
\hline
ReLU	& 85.16\% \\
{\bfseries ELU}	& {\bfseries 88.21\% } \\
SELU	& 87.91\% \\
\hline
\end{tabular}
\end{table}

From the results table, we can observe that ELU achieved the best accuracy on the ResNet-50 model on the CK+ dataset. This stems from the fact that SELU performs much better on models with many layers. In both cases, the change of the activation functions largely outperformed ReLU, which is utilised in most of the modern CNN architectures.

\section{Attention Modules}
Attention modules are designed to be integrated with CNN models to improve them further. 
First, we discuss how attention is implemented in each module, the benefits of each implementation, and possible improvements. Subsequently, we show how the attention modules integrate within the implemented CNN architectures.

\subsubsection{SEN-Net}
The SEN-Net architecture was the first implementation of channel attention in computer vision tasks \cite{Hu2018}. The block improved the representational ability of the network by modelling the interdependencies between the channels of a convolutional layer. This is done through a feature re-calibration operation split into two sequential operations: \textit{squeeze and excitation}.

A set of experiments was conducted on the CK+ dataset using ResNet-50 as the backbone to find the optimal value of $r$ in table~\ref{Sen-net-R}. 

\begin{table}\caption{The Effect of Reduction Ratio Changes on the SEN-Net Attention Module When Applied on the CK+ Dataset}\label{Sen-net-R}
\centering
\begin{tabular}{|c|c|c|}
\hline
Reduction Ratio(r) & \#Parameters & Accuracy \\
\hline
4	& {\bfseries 33.56M} & 87.26\% \\
8	& 28.53M & 88.46\% \\
16	& 26.02M & {\bfseries 89.56\%} \\
32	& 24.76M & 87.91\% \\
\hline
\end{tabular}
\end{table}

\subsubsection{ECA-Net}
ECA-Net \cite{Wang2020} was developed to improve channel attention used in SEN-Net. In SEN-net, the excitation module uses dimensionality reduction via two fully connected layers to extract channel-wise relationships. The channel features are mapped into a low-dimensional space and then mapped back, making the channel connection and weight indirect. Consequently, this negatively affects the direct connections between the channel and its weight, reducing the model's performance. Furthermore, empirical studies show that the operation of dimensional reduction is inefficient and unnecessary for capturing dependencies across all channels \cite{Wang2020}. The ECA-Net attempts to solve the issue of dimensionality reduction while improving the efficiency of the excitation operation by introducing an adaptive kernel size within its excitation operation. 

\begin{equation}
 k= \psi(C)=\left\vert \frac{log_{2}(C)}{\gamma}+\frac{b}{\gamma}\right\vert _{odd}
\end{equation}

A 1D convolutional layer performs the excitation operation with kernel size $k$. The value of $k$ is adaptively changed based on the number of channels. With this operation, ECA captures channel-wise relationships by considering every channel and its $k$ neighbours. Therefore, instead of considering all relationships that may be direct or indirect, an ECA block only considers direct interaction between each channel and its $k$-nearest neighbours to control the model's complexity. Table~\ref{ECA-net-kernel} shows the effect of utilising a static value of $k$ over the adaptive, confirming that the adaptive kernel size is the best option for FER applications.

\begin{table}\caption{The Effect of Kernel Size Changes on the ECA Attention Module When Applied on the CK+ Dataset}\label{ECA-net-kernel}
\centering
\begin{tabular}{|c|c|c|}
\hline
Kernel Size(k) & \#Parameters & Accuracy \\
\hline
1	& 33.56M & 88.46\% \\
3	& 23.50M & 89.65\% \\
5	& 23.53M & 89.11\% \\
7	& 23.59M & 87.36\% \\
9	& 23.62M & 88.56\% \\
Adaptive	& {\bfseries 23.65M } & {\bfseries 90.23\% } \\
\hline
\end{tabular}
\end{table}

\subsubsection{CBAM}
The last attention module implemented in this paper is the Convolutions Block Attention Module (CBAM) \cite{Woo2018}. CBAM proposed utilising both spatial and channel attention to improve the model's performance, unlike the previous attention modules, which only utilised channel attention. The motivation behind the CBAM stemmed from the fact that convolution operations extract informative features by cross-channel and spatial information together. Therefore, emphasising meaningful features along both dimensions should achieve better results.

\textit{CBAM channel attention} consists of squeeze and excitation operations inspired by the implementation of channel attention from SEN-Net\cite{Hu2018}. However, CBAM modifies the original squeeze operation from SEN-net to include average and max pooling to capture channel-wise dependencies. The idea behind utilising both pooling operations stems from the fact that all spatial regions contribute to the average pooling output, whereas max-pooling only considers the maximum values. Consequently, combining both should improve the representation power of relationships between channels.
The two pooling operations are used simultaneously and are passed to a shared network consisting of two fully connected layers ($W_1$ and $W_2$), which perform the excitation operation (following the exact implementation from SEN-Net). After the output of each pooling operation is passed through the shared MLP, the resultant feature vectors are merged using element-wise summation.

The design of the \textit{CBAM spatial} attention module follows the same idea as the \textit{CBAM channel} attention module. To generate a 2D spatial attention map, we compute a 2D spatial descriptor that encodes channel information at each pixel over all spatial locations. This is done via applying average-pooling and max-pooling along the channel axis, after which their outputs are concatenated. This is because pooling along the channel axis effectively detects informative regions as per \cite{Zagoruyko2016}.
The spatial descriptor is then passed to a convolution layer with a kernel size of 7, which outputs the spatial attention map. The choice of the large kernel size is necessary since a large receptive field is usually helpful in deciding spatially important regions. The output is passed through a sigmoid function to normalize the output.

Like SEN-Net, the reduction ratio $r$ allows us to vary the capacity and computational cost of the channel attention block, as shown in a set of experiments that we conducted on the CK+ and summarised in table~\ref{Cbam-R}.
\begin{table}\caption{The Effect of the Change of Reduction Ratio for CBAM Attention Module on the CK+ Dataset}\label{Cbam-R}
\centering
\begin{tabular}{|c|c|c|}
\hline
Reduction Ratio(r) & \#Parameters & Accuracy \\
\hline
4	& {\bfseries 33.57M} & 90.46\% \\
8	& 28.54M & 89.01\% \\
16	& 26.02M & {\bfseries 91.21\%} \\
32	& 24.77M & 90.66\% \\
\hline
\end{tabular}
\end{table}
\subsection{Integration of Different Attention Mechanisms with Different Deep Vision-Based Models}

As mentioned, we integrate the three attention mechanisms discussed earlier with three types of vision-based deep learning architectures. The chosen attention modules are versatile and are designed to be easily integrated within CNN models. 

\subsubsection{Integration with VGGNet} The creators of SEN-Net stated that the SE block could be integrated into standard architectures such as VGGNet by the insertion after the activation layer following each convolution. Through research in the classification of medical images, it was shown that authors had used three different ways to integrate attention in VGGNet: (1) placing attention as described by SEN-Net \cite{Schlemper2019}, (2) placing the attention module before the last fully connected layers \cite{Sitaula2021} and (3) placing the attention modules at layers 11 and 14 \cite{Wang2021-med}. Method 2 achieved the best performance for the emotion recognition task as shown in table~\ref{Vgg-net-attention}. 

\begin{table}\caption{Comparing different attention integration methods for VGGNet when Applied on CK+ Dataset}\label{Vgg-net-attention}
\centering
\begin{tabular}{|c|c|c|}
\hline
Method & \#Parameters & Accuracy \\
\hline
(1)	& {\bfseries  39.99M} & 89.01\% \\
(2)	& 39.95M & {\bfseries 90.11\%} \\
(3)	& 36.81M & 87.91\% \\
\hline
\end{tabular}
\end{table}

\subsubsection{Integration with ResNet}
Even though ResNet is a more complicated architecture, the creators of SEN-Net provided the most optimal way to integrate their block within the residual block, where the attention module is added before summation with the identity branch. Through research and experimentation, we did not find more optimal ways to integrate attention within ResNet; therefore, ECA-Net and CBAM followed the same integration method.

\section{Results}

\subsection{FER Datasets}
It is necessary to have datasets with emotions that are correctly labeled and contain enough data to train the model optimally. For this reason, this paper uses three datasets of different sizes, widely used in FER research.
\textbf{Extended Cohn-Kanade Dataset CK+ dataset}
 \cite{Lucey2010} is an extension of the CK dataset. It contains 593 video sequences and still images of eight facial emotions; Neutral, Angry, Contempt, Disgusted, Fearful, Happy, Sad, and Surprised. The dataset has 123 subjects, and the facial expressions are posed in a lab. The subjects involved are male and female, with a diversity split of 81\% Euro-American, 13\% Afro-American, and 6\% other. 
\textbf{JAFFE Datase}t
\cite{Lyons1998} consists of 213 images of different facial expressions from 10 Japanese female subjects. Each subject was asked to pose seven facial expressions (6 basic and neutral). 
\textbf{FER2013 Dataset}
\cite{Carrier2013} was introduced at the International Conference on Machine Learning (ICML) in 2013 for a Kaggle competition. The training set consists of 28,709 examples, and the public test set consists of 3,589 examples. The samples in the dataset differ in age, race, and facial direction, which closely mimics the real world. The human performance on this dataset is estimated to be 65.5\% \cite{Khaireddin2021}. Hence, it is widely used as a benchmark for emotion recognition models.

\subsection{Evaluation of CNN-Based Models with an ELU Activation Function}
This section shows the results of applying the previously discussed CNN-based models with a different activation function, ELU. This is necessary to establish ground truth and isolate the effect of changing the activation function from adding attention (discussed in the next section). Table~\ref{table1} displays the final evaluation accuracies of the CNN models on the three datasets. The evaluations for CK+ and JAFFE were executed three times to ensure the results' correctness; with smaller datasets, evaluation accuracies fluctuate between the runs. Out of the three executions, the highest value was chosen. 

\begin{table}\caption{Evaluation of CNN architectures with ELU on CK+, JAFFE and FER2013}\label{table1}
\begin{tabular}{|l|l|l|l|l|}
\hline
Architecture &  \#Parameters & CK+ Accuracy  & JAFFE Accuracy  & FER2013 Accuracy\\
\hline
VGG-16	& 39.92M	& 87.91\%	& 64.44\%	& 60.66\% \\
VGG-19	& {\bfseries 42.87M}	& {\bfseries 90.66\%}	& {\bfseries 68.89\%}	& {\bfseries 60.92\% } \\
\hline
ResNet-50 & 23.49M & 87.91\% & 	{\bfseries 73.33\%}	& 58.61\% \\
ResNet-101	&  42.46M & {\bfseries 88.46\%}	& 60.00\%	& 58.67\% \\
ResNet-152	& {\bfseries 58.08M} & 85.71\%	& 15.66\%	&  {\bfseries 59.36\%} \\
\hline
ResNetV2-50 & 23.48M & 	88.46\%	& {\bfseries 77.78\%}	&  58.72\% \\
ResNetV2-101 & 	42.44M	& 88.62\%	&  62.22\%	& 59.07\% \\
ResNetV2-152 & {\bfseries 58.05M} & {\bfseries 89.01\%} & 66.67\% & {\bfseries 59.40\%} \\
\hline
\end{tabular}
\end{table}

Analysing the results, we see that VGG-19 achieved the best accuracy on CK+ and FER2013, while ResNetV2-50 achieved the best accuracy on the JAFFE dataset. This was an unexpected result as the initial assumption was that the deeper ResNet models should outperform VGGNet, which was not the case. We conclude that this is due to the modification of the activation function from ReLU to ELU in the CNN models. This change improved the VGG-19 accuracy from 87.91\% to 90.66\% on the CK+ dataset, significantly better than the deeper ResNet models for the same modification. This finding indicates that residual learning is not required to achieve good performance. Even simple architectures such as the VGGNet can achieve higher accuracy than a more complex architecture such as ResNet across different datasets by utilising the ELU activation functions.
Furthermore, the deeper ResNet models consist of more parameters than VGG-19. Because deep CNNs are designed to be trained on large amounts of data, the layers at the deeper stages cannot learn informative features. Consequently, overfitting occurs, suggesting that shallower architectures are better for the given dataset. It is yet to be discovered whether a larger dataset would enhance the performance of the deeper architecture of ResNet. 

\begin{figure}
\includegraphics[width=1.05\textwidth]{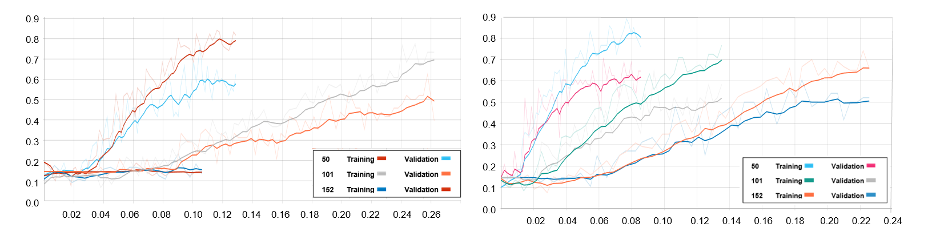}
\caption{raining and validation accuracy graphs of ResNet(left) and ResNetV2(right) with 3 Different Depths (50, 101 and 152) on the JAFFE dataset.} \label{Resultsfig1}
\end{figure}

From the previous table, it can be seen that ResNet performed better than VGG on the smaller JAFFE dataset. To gain further insight into the baseline performances of the two ResNet architectures, we drill down more by comparing the relative training graphs of ResNetV1 and ResNetV2 on the JAFFE dataset in Figure~\ref{Resultsfig1}. Interestingly, the figures show that ResNetV2 performed significantly better than ResNet on the smallest JAFFE dataset. Original ResNet showed degradation in accuracy past depth 101 and could not increase training accuracy past depth 152 on the JAFFE dataset. On the other hand, ResNetV2 can still train on the deeper models, and the model of 50 layers performed better than the original ResNet.

Relative graphs were chosen to separate the ResNet models as the larger models will have a longer computational time. Figure~\ref{Resultsfig1} shows that ResNetV2 converges to optimal values faster, and the performance degradation in the deeper layers is not as sudden as the original ResNet.
From these results, we can conclude that the ELU activation function further enhanced the new residual blocks due to its ability to facilitate a better flow of information. This, however, should not be attributed only to the small size of the JAFFE dataset since the new improved residual blocks also performed consistently better on CK+ and FER2013 datasets.

\subsection{CNNs with Different Attention Mechanisms}
This section shows the results of augmenting the previously discussed CNN-based architectures with different attention mechanisms.

\begin{table}
\caption{Evaluation of SEN-Net, ECA-Net and CBAM Attention Modules when Infused in VGG, ResNet and ResNetV2 with Different Depths, with ELU Activation Function, Applied on CK+, JAFFE, and FER2013 }\label{table2}
\begin{tabular}{|l|l|l|l|l|}
\hline
Architecture &  Param & CK+ Accuracy  & JAFFE Accuracy  & FER2013 Accuracy\\
\hline
VGG-16	& 39.92 M & 87.91\%	 & 64.44\%	 & 60.66\% \\
VGG-16 + SEN-Net & 39.95M & 88.46\% & 68.89\% & 63.05\% \\
VGG-16 + ECA-Net & 39.92M & 89.01\% & 73.33\% & 62.72\% \\
VGG-16 + CBAM & {\bfseries 39.95M} & {\bfseries 89.56\%} & {\bfseries 75.56\%} & {\bfseries 63.46\%} \\
\hline
VGG-19 & 42.87M & 90.66\% & 68.89\% & 60.92\% \\
VGG-19 + SEN-Net & 45.26M & 91.21\% & 73.33\% & 63.23\% \\
VGG-19 + ECA-Net & 45.23M & 91.76\% & 75.56\% & 63.49\% \\
VGG-19 + CBAM & {\bfseries 45.26M} & {\bfseries 92.31\% (↑ 1.65\%)} & {\bfseries 77.78\%} & {\bfseries 64.07\% (↑ 3.15\%)} \\
\hline
ResNet-50 & 23.49M & 87.91\% & 73.33\% & 58.61\% \\
ResNet-50 + SEN-Net & 26.02M & 89.01\% & 75.56\% & 58.84\% \\
ResNet-50 + ECA-Net & 23.65M & 90.11\% & 77.78\% & 59.73\% \\
ResNet-50 + CBAM & {\bfseries 26.02M} & {\bfseries91.21\%} & {\bfseries 82.22\%} & {\bfseries 59.90\%} \\
\hline
ResNet-101 & 42.46M & 88.46\% & 60.00\% & 58.67\% \\
ResNet-101 + SEN-Net & 47.24M & 89.01\% & 68.89\% & 58.92\% \\
ResNet-101 + ECA-Net & 42.81M & 89.56\% & 73.33\% & 60.15\% \\
ResNet-101 + CBAM & {\bfseries 47.24M} & {\bfseries 90.11\%} & {\bfseries 75.56\%} & {\bfseries 60.92\%} \\
\hline
ResNet-152 & 58.08M & 85.71\% & 15.66\% & 59.36\% \\
ResNet-152 + SEN-Net & 64,71M & 88.46\%	 & 15.66\%	 & 59.73\% \\
ResNet-152 + ECA-Net & 58.60M & 89.56\%	 & 15.66\%	 & 60.92\% \\
ResNet-152 + CBAM	& {\bfseries 64.71M}	& {\bfseries 90.11\%}	& {\bfseries 15.66\%}	& {\bfseries 61.54\%} \\
\hline
ResNetV2-50	& 23.48M	& 	88.46\%	& 	77.78\%	& 	58.72\% \\
ResNetV2-50 + SEN-Net & 26.01M & 88.66\% & 82.22\% & 59.36\% \\
ResNetV2-50 + ECA-Net	& 	23.64M	& 	88.91\%	 & 82.22\% 	& 59.73\% \\
ResNetV2-50 + CBAM	 &{\bfseries  26.01M}	& {\bfseries 89.01\% }	& {\bfseries 84.44\%(↑ 6.55\%)} & {\bfseries 60.15\%} \\
\hline
ResNetV2-101 & 42.44M	& 88.62\%	& 62.22\%	& 59.07\% \\
ResNetV2-101 + SEN-Net	& 47,22M	& 89.01\%	& 68.89\%	& 59.73\% \\
ResNetV2-101 + ECA-Net	& 42.79M	& 89.56\%	& 70.83\%	& 60.15\% \\
ResNetV2-101 + CBAM	& {\bfseries 47.22M}	& {\bfseries 90.66\%}	& {\bfseries 73.33\%}	& {\bfseries 60.92\%} \\
\hline
ResNetV2-152	& 58.05M	& 89.01\%	& 66.67\%	& 59.40\% \\
ResNetV2-152 + SEN-Net	& 64.68M	& 89.56\%	& 68.89\%	& 60.72\% \\
ResNetV2-152 + ECA-Net	& 58.57M	& 89.82\%	& 73.33\%	& 61.54\% \\
ResNetV2-152 + CBAM	& {\bfseries 64.69M}	& {\bfseries 90.11\%}	& {\bfseries 77.78\%}	& {\bfseries 62.05\%} \\
\hline
\end{tabular}
\end{table}
Table~\ref{table2} summarizes the experimental results. The networks with attention outperformed all the baselines significantly, demonstrating that attention can generalise well on various models. Moreover, the addition of attention showed performance improvement across the three studied datasets, displaying that attention could be applied to any problem size.

\begin{figure}
\includegraphics[width=\textwidth]{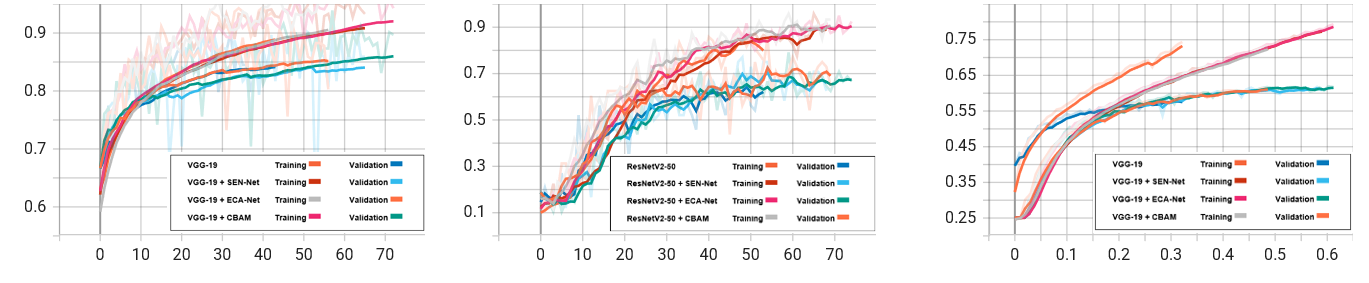}
\caption{Accuracy curves for the best performing models on the CK+(left), JAFFE(middle), and FER2013(right). } \label{Resultsfig3}
\end{figure}

Figure~\ref{Resultsfig3} shows the accuracy curves of the best-performing networks. In each case, attention achieves higher accuracies and shows a smaller gap between training and validation curves than baseline networks. 

As expected, CBAM had the best improvement in accuracy over the other attention modules due to the application of spatial attention. However, that comes at the cost of a significant overhead in parameters. On the other hand, ECA-Net achieved similar levels of performance increase compared to CBAM while not significantly impacting the memory requirement of each network. 
VGG19 still achieved the best performance on the CK+ and FER2013 datasets, while ResNetV2-50 achieved the best performance on the JAFFE dataset. However, the increase in performance was significantly higher than expected in the FER2013 dataset. Due to the size of the dataset, the expected improvement should have been 1-2\% which is the improvement authors of CBAM received on the ImageNet dataset. However, CBAM achieved a performance increase of 3.15\% on FER2013, displaying that attention modules can significantly impact the network's performance. Furthermore, the addition of CBAM enabled an increase of 6.55\% on the JAFFE dataset, demonstrating the ability of attention modules to improve the network's generalisation ability. Additionally, the introduction of attention did not change the ranking order of the best-performing networks from the baseline CNN comparisons, emphasising the consistency of the expected boost in performance when the attention mechanism is added.

\section{Conclusion}
In this paper, we studied the effect of infusing three different attention mechanisms, SEN-Net, ECA-Net, and CBAM, into three CNN-based deep learning architectures, namely the VGGNet, ResNet, and ResNetV2, with different depths to classify the seven basic human emotions on three datasets, namely CK+, JAFFE, and FER2013. In addition, we have replaced their internal activation function from RELU to ELU. As a result, there was a significant improvement in their performances. We studied the effect of changing the activation function first, then infused the resultant architectures with attention. We also showed that the new residual blocks presented in ResNetV2 perform significantly better than the original ResNet on smaller datasets and slightly improve on mid-sized and larger-sized datasets. 
Our results show that these amendments refined the extracted features and improved the generalisation capabilities of these models. The attention module hyperparameters were modified through experimentation to maximize the models' performance on emotion recognition tasks. 

Our work verified the attention mechanism's effect on the performance of CNNs. We have shown that each attention module outperformed the baseline models on each dataset. Consequently, attention modules could successfully improve the generalisation ability and refine the extracted features regardless of the problem size. 
Furthermore, our work confirmed that utilising ResNet V2 with attention modules yields better results than the original ResNet when attention modules and ELU are applied. In the future, we intend to conduct a comprehensive study on the effect of simplifying the transformation operations used in attention to speed up training time without losing competency.

\bibliographystyle{splncs04}
\bibliography{paper}
\end{document}